\documentclass{article}

\usepackage[preprint, nonatbib]{neurips_2023}

\usepackage[utf8]{inputenc} 
\usepackage[T1]{fontenc}    
\usepackage{hyperref}       
\usepackage{url}            
\usepackage{booktabs}       
\usepackage{amsfonts}       
\usepackage{nicefrac}       
\usepackage{microtype}      
\usepackage{xcolor}         
\usepackage{graphicx}       
\usepackage{multirow}
\usepackage{tcolorbox}
\usepackage{amsmath}

\title{From Simple to Professional: A Combinatorial Controllable Image Captioning Agent}

\author{%
  Xinran Wang \\
  PRIS Lab, BUPT\\
  \texttt{wangxr@bupt.edu.cn} \\
  \And
  Muxi Diao \\
  PRIS Lab, BUPT \\
  \texttt{dmx@bupt.edu.cn} \\
  \And
  Baoteng Li \\
  PRIS Lab, BUPT \\
  \texttt{meltry.lbt@bupt.edu.cn} \\
  \And
  Haiwen Zhang \\
  PRIS Lab, BUPT \\
  \texttt{zhanghaiwen@bupt.edu.cn} \\
  \AND
  Kongming Liang\thanks{Corresponding author} \\
  PRIS Lab, BUPT \\
  \texttt{liangkongming@bupt.edu.cn} \\
  \AND
  Zhanyu Ma \\
  PRIS Lab, BUPT \\
  \texttt{mazhanyu@bupt.edu.cn} \\
}

\begin{document}

\maketitle

\begin{abstract}
  The Controllable Image Captioning Agent (CapAgent) is an innovative system designed to bridge the gap between user simplicity and professional-level outputs in image captioning tasks. CapAgent automatically transforms user-provided simple instructions into detailed, professional instructions, enabling precise and context-aware caption generation. By leveraging multimodal large language models (MLLMs) and external tools such as object detection tool and search engines, the system ensures that captions adhere to specified guidelines, including sentiment, keywords, focus, and formatting. CapAgent transparently controls each step of the captioning process, and showcases its reasoning and tool usage at every step, fostering user trust and engagement. The project code is available at \url{https://github.com/xin-ran-w/CapAgent}.
\end{abstract}

\section{Introduction}

\begin{figure}
    \centering
    \includegraphics[width=\linewidth]{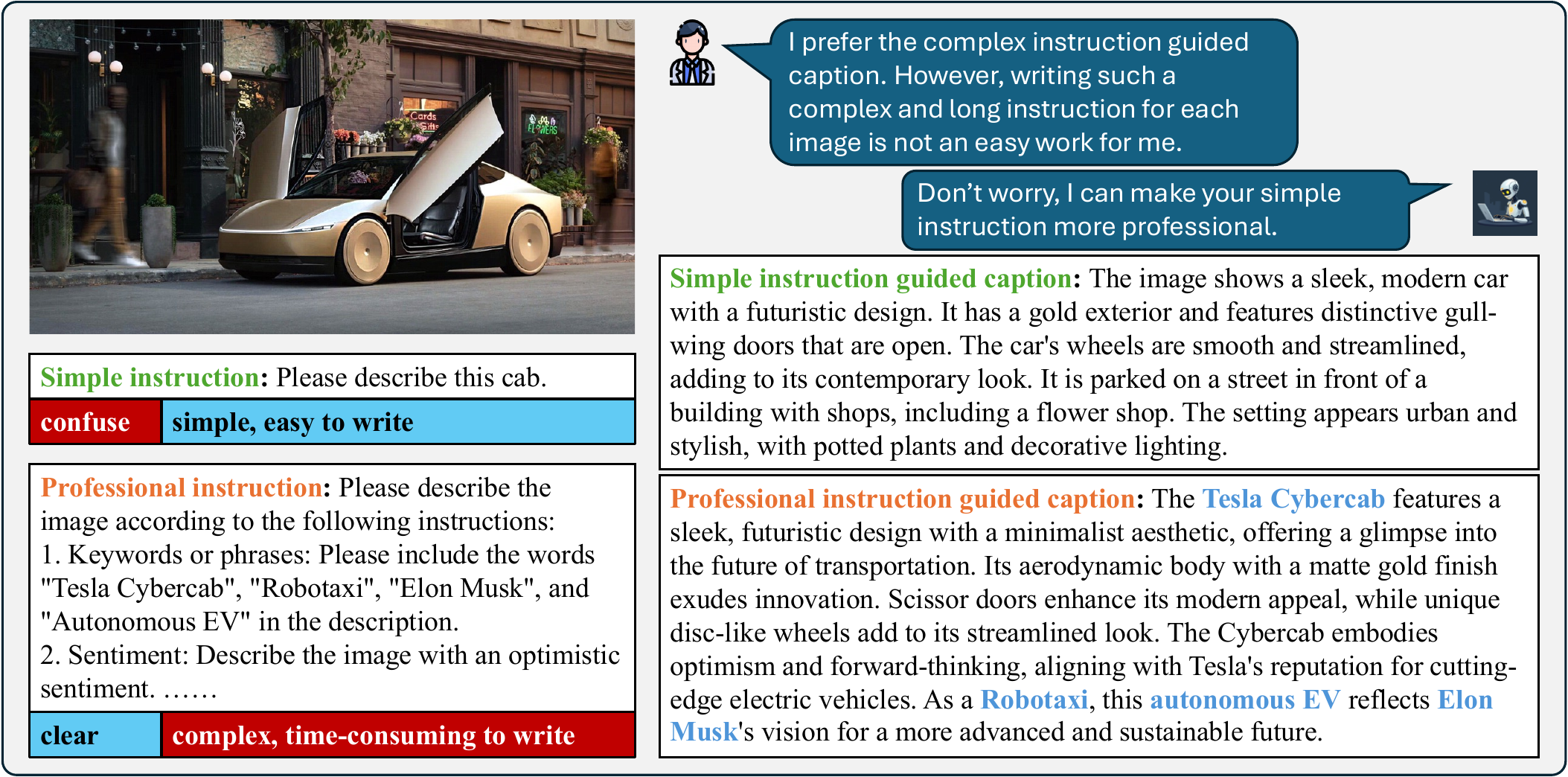}
    \caption{When a user needs to describe an image, he/she may need to write more complex and professional instructions to achieve the effect he/she wants. However, writing such long and complex instructions is not easy for the user. Our CapAgent can automatically make the user instructions more professional and follow the evolved instructions to generate more professional captions.}
    \label{fig:intro}
\end{figure}

Image captioning \cite{dwibedi-2024-arXiv-flexcap, chen-2015-axXiv-cococap, onoe-2024-arXiv-docci, dong-2024-arXiv-detailcaps, biten-2019-CVPR-goodnews, garg-2024-arXiv-imageinwords, sharma-2018-ACL-conceptual, mathews-2016-AAAI-senticap, tran-2020-NYTimes800k, wang-2023-arXiv-captionanything}, a pivotal task in computer vision and natural language processing, aims to articulate the visual content of an image through text. It underpins a wide range of applications, including assistive technologies, content creation, and image understanding.

With the emergence of visual language models (VLMs), the ability to generate detailed image descriptions has seen significant advancements. Large multimodal models, such as GPT-4o \cite{openai-2024-gpt4o-card}, can produce richly detailed captions with a simple prompt, e.g., ``Please describe this image in detail.''. However, such generic prompts lack constraints, often leading to captions that may not align with user preferences. One approach to addressing this is by crafting professional instruction with complex combinatorial constraints, e.g., ``within 80 words in Chinese'', ``if the image contains more than three objects using bullet format to describe each object'', to control the generated caption. While effective, creating such tailored instructions is time-consuming and cumbersome for general users. Additionally, these instructions often need to be customized based on the specific visual content of each image, adding another layer of complexity. To simplify this process, we apply instruction evolving to transform simple user instructions into context-aware professional instructions tailored to the visual content of the corresponding images.

However, even with professional instructions, current large multimodal models face challenges in achieving fine-grained control over caption generation, particularly when handling combinatorial constraints across multiple descriptive aspects. To address this, we introduce \textbf{CapAgent}, an agent-based system equipped with a suite of tools designed to transform static caption generation into a dynamically controllable process. CapAgent empowers users to specify and enforce multiple types of constraints, as defined in prior work \cite{wen-2024-arXiv-benchmarking-complex-if}, such as: (1) \textbf{Format constraint} (e.g., caption length, caption format); (2) \textbf{Semantic constraint} (e.g., user-defined interests, important details of key objects, caption sentiment); (3) \textbf{Lexical constraint} (e.g., captioning the image with input keywords); (4) \textbf{Utility constraint} (e.g., sentiment, lexical choices).

\section{Method}

\subsection{Overview}

Our method has two steps. The first step, instruction evolving, helps users convert simple instructions into professional instructions. The second step is to apply an agent system, named CapAgent, to generate captions that follow professional instructions by using various tools. We'll go into the details of these two steps in the following section.

\begin{table}[htbp]
    
    \caption{Constraint dimensions for evolving image captioning instructions.}
    \begin{tcolorbox}
    
\textbf{View}: The view to describe the image, e.g., from the view of the middle object. \\

\textbf{Sentiment}: Captures the emotional tone of the image, ensuring the description reflects its mood or atmosphere. \\

\textbf{Focus content}: Emphasizes the inclusion of key details about salient objects in the image. \\

\textbf{Keywords}: Ensures the caption includes specific predefined keywords. \\

\textbf{Length}: The length of the output sentence, e.g., using two sentences. \\

\textbf{Format}: Some images have strong emotions, and it is a better choice to describe the image with the corresponding emotional tone. \\

\textbf{Genre}: The genre of the image caption, for example, poetry, travel blog or Instagram post.

    \end{tcolorbox}
    \label{tab:constraint-dim}
\end{table}

\subsection{Instruction Evolving}

When interacting with chatbots, general users tend to prefer providing simple instructions over complex ones \cite{wen-2024-arXiv-benchmarking-complex-if}. However, simple instructions often lack sufficient detail, leaving the model uncertain about the user's exact needs. While complex instructions with combinatorial constraints can better articulate the user's intent, enabling the model to generate outputs more aligned with their expectations, crafting such detailed instructions is often time-consuming and burdensome for general users.

To improve the user experience, we introduced a module designed to transform simple user inputs into well-structured, complex professional instructions. Formally, given a user input caption instruction $s$ and its corresponding image $x$, an evolved instruction $s^\star$ can be obtained by $A(s,x)$, where $A$ is an instruction evolving model, such as a multimodal large language model (MLLM). There are three main challenges for instruction evolution: \textbf{(1) Direction of evolving}: Which dimensions of the caption can be customized? \textbf{(2) Contextual integration}: How to combine the context information of the image effectively? \textbf{(3) Evolving criteria}: What criteria should the evolved instruction meet to ensure quality and relevance?

\paragraph{Direction of evolving} Inspired by controllable text generation, we select several important constraint types suitable for caption customization listed in Table \ref{tab:constraint-dim}. Each type of constraint is responsible for controlling different dimensions of the caption.

\begin{figure}
    \centering
    \includegraphics[width=\linewidth]{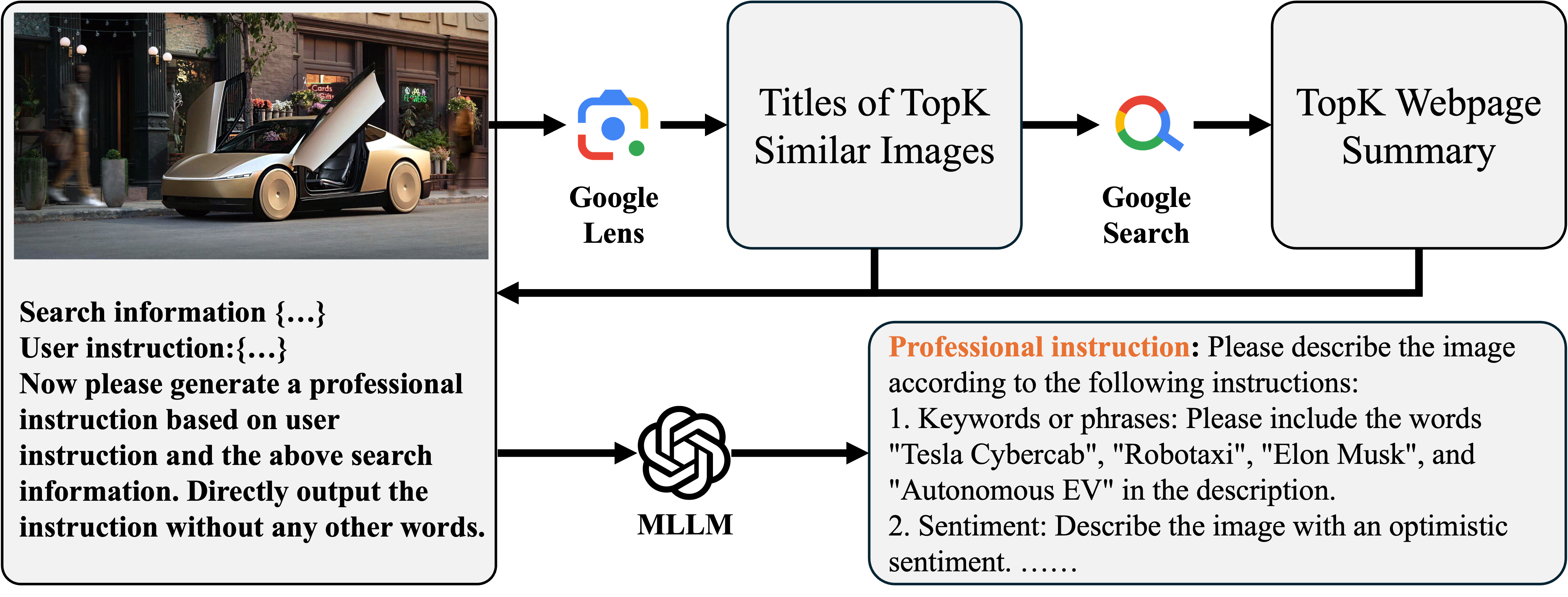}
    \caption{Process of instruction evolution with context information extracted from the webpage by using Google Lens and Google Search.}
    \label{fig:search}
\end{figure}

\paragraph{Evolving with context} Some images contain additional context information that is not immediately apparent. For example, the car in Figure \ref{fig:intro} is a well-known product, the "Tesla Cybercab." The instruction evolution model alone may not accurately identify such context by analyzing the image in isolation. To address this limitation, we enable the instruction evolution model to incorporate external context by leveraging web information through APIs such as Google Search\footnote{https://www.google.com} for text search and Google Lens\footnote{https://lens.google} for image search. The process of integrating web-accessed context is illustrated in Figure \ref{fig:search}. We first use Google Lens to search top $K$ similar images on the web. Then, we use the LLM to generate a summary of the phrases or sentences that need to be further searched according to titles of similar images. Finally, we use the LLM summary of the Google Search results and Google Lens results as the image context $c$. We combine user instruction $s$ and image $x$ with $c$ together to prompt MLLM to generate the professional instruction, which is formulated as $s^* = A(s, c, x)$.

\paragraph{Evolving criteria} The evolved instruction $s^\star$ is expected to satisfy the following criteria:
\begin{itemize}

    \item \textbf{Correctness}: The professional instructions need to be describable by humans. For example, describing objects that do not exist in the image makes the instruction unanswerable. \cite{qian-2024-arXiv-mia}

    \item \textbf{Content Specificity}: The augmented instruction $s^\star$ should be tailored according to the visual content of $x$.
    
    \item \textbf{Consistency}: The augmented instruction $s^\star$ should maintain logical coherence and avoid contradictions.

    \item \textbf{Constraint Inheritance}: Any explicit constraints or preferences stated in the user's instruction $s$ must be preserved and appropriately integrated into the augmented instruction $s^\star$.
\end{itemize}

We employ GPT-4o as the instruction evolving module. After the instruction evolved, we input the evolved instruction $s^\star$ to an image captioning model to gain the image caption.



\subsection{CapAgent}
\begin{figure}[h]
    \centering
    \includegraphics[width=\linewidth]{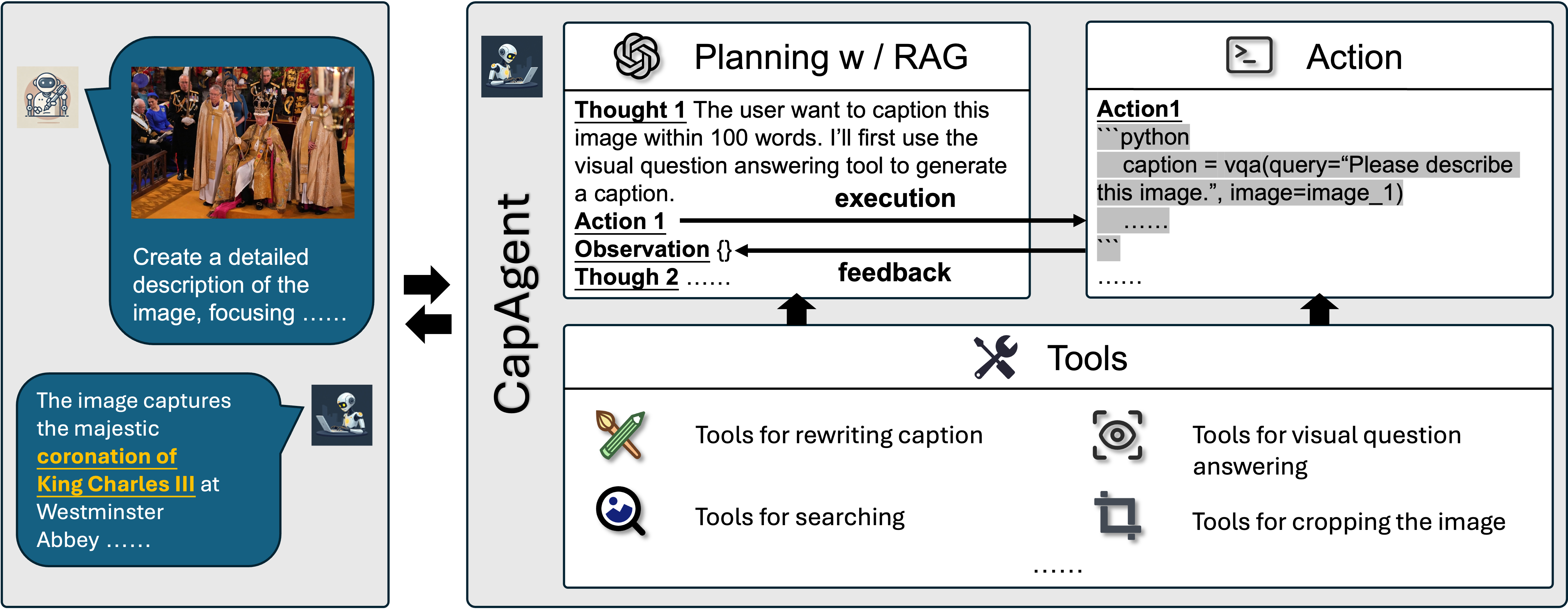}
    \caption{The diagram of CapAgent's workflow. }
    \label{fig:method}
\end{figure}

Although professional instruction may express user intent, MLLM may not generate a caption fully following the professional instruction.
In this section, we introduce the CapAgent, an agent system with a variety of tools specifically designed to control the image captioning process.
As shown in Figure \ref{fig:method}, like most general agents, CapAgent's workflow includes three main steps: planning, tool usage, and observation. When the user inputs an image and a caption query, the CapAgent will generate a thought and a corresponding action to tackle the user request. In each action, the agent will generate a block of Python code and execute the code to observe the action result.

\subsubsection{Thought, Action and Observation}
Verbal reasoning is a fundamental ability of human intelligence \cite{hu-2024-NIPS-visual-sketchpad}. Given user instruction $s$, our agent will generate the initial thought and action, $t_0$ and $a_0$. The action is a piece of Python code and it will be executed on a local Jupyter server. The execution result will be fed back to the CapAgent to generate the thought and action for the next step until all user requirements are met.
Here, we use the ReAct \cite{yao-ICLR-2023-react} prompt to let the planner generate thought and action for each step.



\subsubsection{Retrieval Augmented Planning}

To enhance the precision of thought and action generation within our model, we have curated a set of thought, action, and observation chain examples to guide CapAgent in adapting to the task and producing outputs in an appropriate format. Given the limited context capacity of MLLM, we cannot embed all examples within the prompt, necessitating a selection process. We hypothesize that similar user demands should elicit similar thought processes from the agent. To this end, we employ Retrieval-Augmented Generation (RAG) \cite{lewis-2020-NIPS-RAG, kagaya-2024-arXiv-RAP} to identify the top $N$ most analogous examples based on the user's specific requirements.

Formally, given a user instruction $s$ and a database of chain-of-thought examples $C=\left\{c_i\right\}_{i=1}^T$, our objective is to extract a subset $C_s=\left\{c_i\right\}_{i=1}^N$ of size $N$ from $C$ that closely matches the user's specific requirements. We achieve this by first applying the document embedding model BGE-M3 \cite{chen-2024-arXiv-bgem3} to embed each $c_i$ into a vector database, transforming each chain-of-thought example into a vector representation $v_i$. The user's instruction $s$ is also embedded into the same vector space, resulting in $v_s$. The similarity between the user's instruction embedding $v_s$ and each thought chain example embedding $v_i$ is calculated using cosine similarity:

\begin{equation}
    \operatorname{sim}\left(\mathbf{v}_s, \mathbf{v}_i\right)=\frac{\mathbf{v}_s \cdot \mathbf{v}_i}{\left\|\mathbf{v}_s\right\| \left\|\mathbf{v}_i \right\|}
\end{equation}

The top $N$ thought chain examples with the highest similarity scores are selected to form the subset $C_s$. This can be represented as:

\begin{equation}
C_s=\operatorname{TopN}\left(\left\{c_i \mid \operatorname{sim}\left(v_s, v_i\right)\right\}_{i=1}^{T}\right)
\end{equation}

After that, we put $C_s$ into the end of the system prompt to enable CapAgent to generate the correct thoughts and actions.

\subsubsection{Tools for Combinatorial Caption Control}

We have developed a comprehensive suite of tools to facilitate CapAgent's combinatorial control over the captioning process. These tools leverage the APIs of foundational models, including GPT-4o \cite{openai-2024-gpt4o-card}, GroundingDINO \cite{liu-2024-arXiv-groundingdino}, and DepthAnything v2 \cite{yang-2024-depth-v2}.

\paragraph{Visual question answering tool} This tool is instrumental in generating basic captions and responding to queries about objects within an image. It serves as the initial step in providing a descriptive narrative of the visual content.

\paragraph{Caption sentiment modification tool} This tool allows for the adjustment of caption sentiment. By providing an initial caption and a desired sentiment, it generates a new caption that aligns with the specified emotional tone while retaining the original content.

\paragraph{Caption expansion tool} When the length of the basic caption is shorter than the user requirement, CapAgent can use this tool to extend the caption content by asking and answering questions about the image until it meets the required caption length.

\paragraph{Caption condensation tool} Conversely, when the initial caption exceeds the specified length, this tool is used to condense the caption. It removes superfluous details, ensuring the caption remains concise and relevant.

\paragraph{Object counting tool}
Designed to cater to the need for quantification, this tool enables CapAgent to count specific objects within an image. It is particularly useful when users require a numerical description of elements within the visual content.

\paragraph{Spatial relation tool} This tool is crucial for describing the spatial layout of images, especially when the content is confined to a specific space, such as a living room. It is especially invaluable for providing environmental context to vision-impaired individuals. Enabled by object detection and depth estimation vision expert models, it accurately portrays the spatial relationships within the image.

\section{Visualization}

\begin{figure}[htbp]
    \centering
    \includegraphics[width=\linewidth]{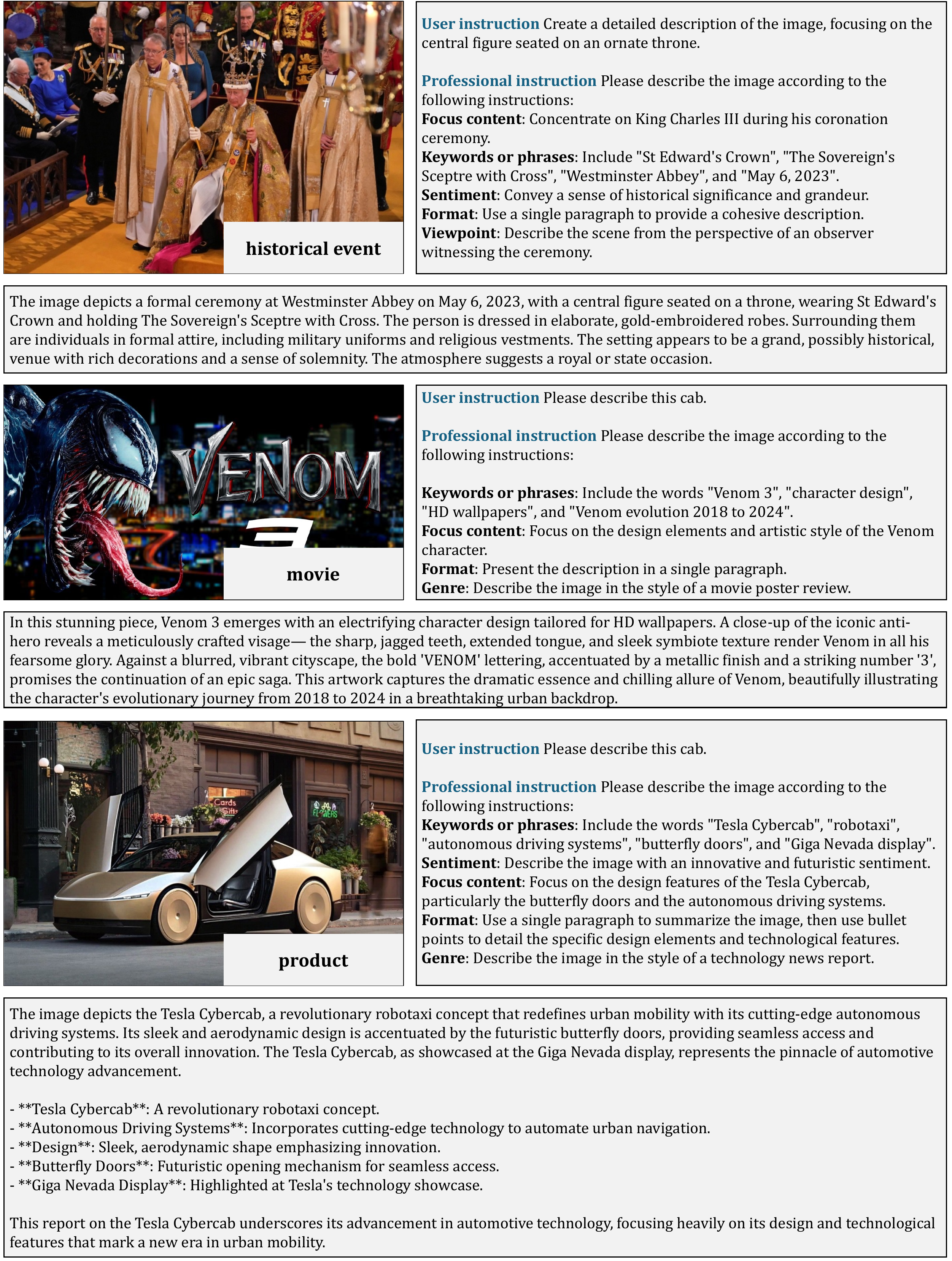}
    \caption{Professional instruction and caption examples of CapAgent (Page 1).}
    \label{fig:vis_demo_1}
\end{figure}

\begin{figure}[htbp]
    \centering
    \includegraphics[width=\linewidth]{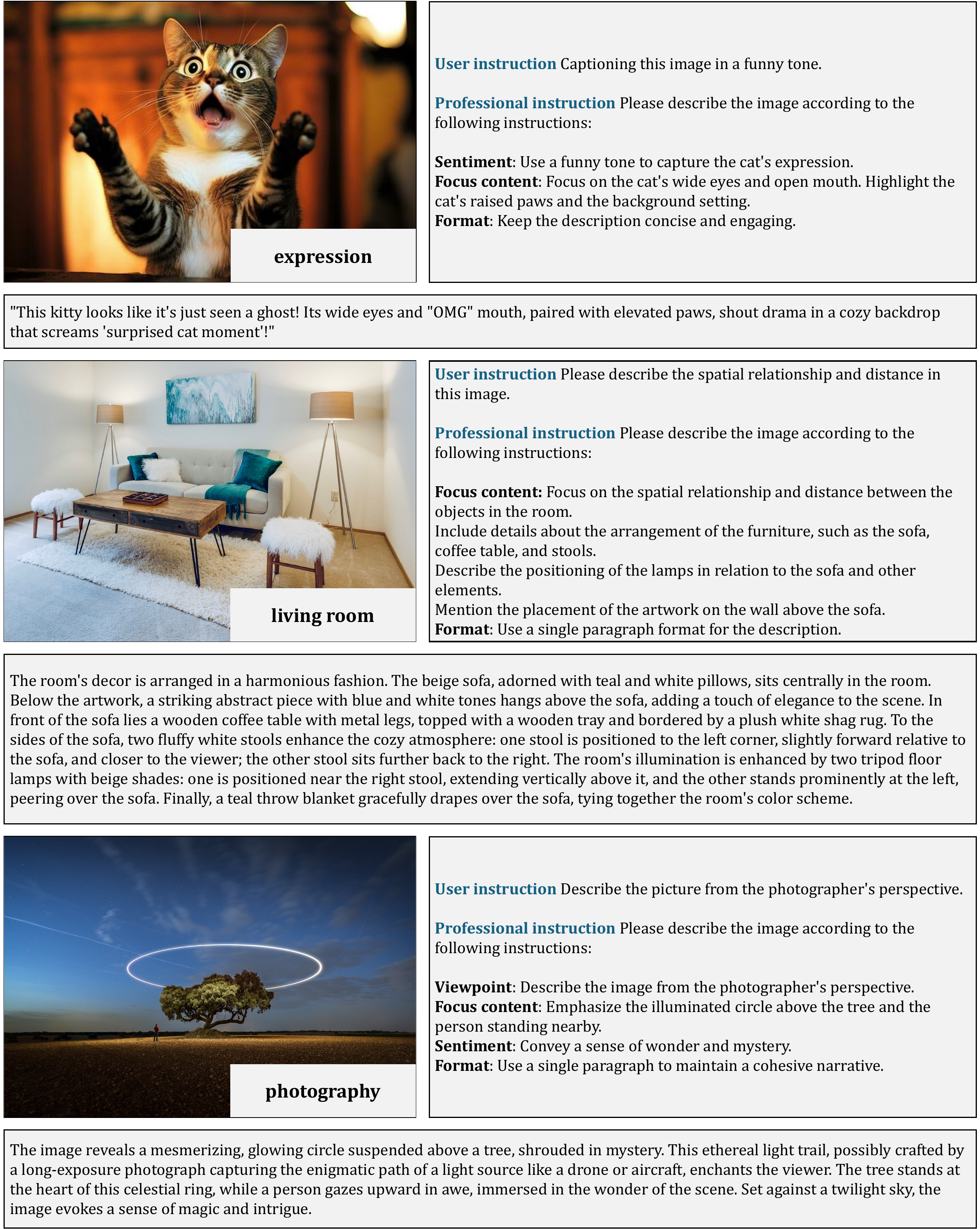}
    \caption{Professional instruction and caption examples of CapAgent (Page 2).}
    \label{fig:vis_demo_2}
\end{figure}

The visualization examples in Figure \ref{fig:vis_demo_1} and Figure \ref{fig:vis_demo_2} demonstrate the model's ability to generate detailed and contextually rich captions for a variety of image categories, including historical events, movie posters, and other topics. Additionally, the professional instructions generated by CapAgent are tailored to meet the specific requirements based on the content of the image and the user's input. For instance, when processing a movie poster, the professional instruction explicitly specifies that the final caption should include keywords such as ``Venom 3'' and ``HD wallpapers'', ensuring relevance to both the visual content and the context of the image. These captions not only describe the scenes but also capture the sentiment and key elements outlined in the professional instructions. As a result, they align with the user's expectations and provide a comprehensive visualization of the depicted events and products.

\section{Conclusion}

In conclusion, CapAgent showcases a powerful capability to bridge the gap between user-provided simple instructions and professional-grade image captions. By leveraging advanced tool integration and professional instruction generation, the system delivers contextually accurate, sentiment-aligned, and detail-rich descriptions for a wide range of images. This ensures a seamless alignment with user intent while maintaining precision and contextual relevance.

\bibliographystyle{plain} 
\bibliography{ref} 

\end{document}